\documentclass{article}

\PassOptionsToPackage{numbers,sort&compress}{natbib}
\usepackage[dblblindworkshop, final]{neurips_2026}
\workshoptitle{Neural Network Artifacts as a New Data Modality}
\usepackage[utf8]{inputenc}
\usepackage[T1]{fontenc}
\usepackage{hyperref}
\usepackage{url}
\usepackage{booktabs}
\usepackage{graphicx}
\usepackage{amsfonts}
\usepackage{amsmath}
\usepackage{multirow}
\usepackage{nicefrac}
\usepackage{microtype}
\usepackage{xcolor}
\usepackage{tabularx}
\usepackage{placeins}


\renewcommand{\topfraction}{0.9}
\renewcommand{\bottomfraction}{0.6}
\renewcommand{\textfraction}{0.1}
\renewcommand{\floatpagefraction}{0.75}
\newcolumntype{L}{>{\raggedright\arraybackslash}X}
\newcolumntype{R}{>{\raggedleft\arraybackslash}X}

\title{Model Merging Does Not Require Uniform LoRA Ranks}

\author{%
  Xie Zirong \\
  Amazon \\
  \texttt{zirongx@amazon.co.jp}
  \And
  Zainab Afolabi  \\
  Amazon Web Services \\
  \texttt{zafolabi@amazon.com} \\
  \And
  Nikita Kozodoi\\
  Amazon Web Services \\
  \texttt{kozodoi@amazon.com} \\
  \And
  Jack Butler \\
  Amazon Web Services \\
  \texttt{jackbtlr@amazon.com} \\
}

\begin{document}

\maketitle
\begin{abstract}
Model merging combines separately fine-tuned experts into a single model without retraining. Standard practice trains every expert at the same LoRA rank, which couples the rank choice across experts even though each domain has its own best rank. We show this coupling is a property of the merge method, not of merging. We train 36 experts on three domains (mathematics, instruction following, multilingual) at four ranks and three model scales, then merge with six methods at uniform ranks and at each domain's own best rank. Giving experts different ranks costs averaging and utility-based selection nothing, and gains them up to 1.8 pp of multi-domain accuracy, while the conflict-resolution methods are inconsistent and can lose up to 4 pp. The split has a simple cause. Merging keeps only a fixed share of the pooled spectral components, and methods that choose them by size discard many that would have reduced loss. They also tend to discard the same expert: the instruction adapter has the smallest singular values, so it is dropped almost entirely and IFEval falls by 16 to 25 pp. Choosing components by their effect on loss instead keeps all three experts, and leaves accuracy nearly flat across an 8$\times$ rank range (0.7 pp, against 6.7 pp for TIES at 4B scale). Rank can therefore be chosen per expert whenever the merge operator does not select by magnitude.
\end{abstract}

\section{Introduction}
\label{sec:intro}

Post-training on specialized domains produces experts that excel individually but cannot be deployed together without multiplying inference cost. Model merging is the post-training step that resolves this, combining the experts into one model that handles all tasks. The technique has matured rapidly: from model soups~\citep{wortsman2022soups} to TIES~\citep{yadav2023ties}, DARE~\citep{yu2024dare} and KnOTS~\citep{stoica2024knots}, each method improves on simple averaging, and toolkits like MergeKit~\citep{goddard2024arcee} make experimentation accessible. Yet across this line of work the LoRA rank is fixed uniformly across experts, and merging adapters of differing rank is largely unexplored.

This convention has a cost. Rank sets adapter capacity, and how much is needed varies with the task, since domains differ in how far they shift the base model and in how much training data they have. A shared rank therefore forces at least one expert away from the rank that suits it. Rank is fixed during fine-tuning, yet the constraint on it comes from the later merging step, so the two stages cannot be designed independently. We measure the per-domain optima in Section~\ref{sec:setup} and then ask whether merging really requires this coupling at all.

The answer depends on the merge method, and the reason is structural. When experts share rank $r$, the truncation budget $R = \max_i r_i$ admits a fixed fraction of the pooled spectrum however large $r$ is. Diverse ranks relax that budget, but magnitude-based selection cannot use the extra room, because it keeps the largest singular values whether or not they reduce loss. We formalize a first-order utility criterion and find that most of the components it rejects would have reduced loss, so the operator discards useful directions by construction rather than by bad tuning.

We make two contributions. First, we show that giving experts different ranks costs averaging and utility-based selection nothing and gains them up to 1.8 pp, while conflict-resolution methods such as TIES can lose up to 4 pp with inconsistent performance, and we trace that asymmetry to how each operator picks spectral components. Second, we identify what the failure destroys rather than only how much: magnitude truncation pushes a single expert out of the kept subspace, and the loss lands almost entirely on instruction following. Together these say that rank coordination across experts is a constraint of the merge method, so changing the method removes it.

\section{Related Work}
\label{sec:related}

Model merging began with simple averaging~\citep{wortsman2022soups} and task arithmetic~\citep{ilharco2023taskarithmetic}, which scales and adds task vectors. These work well when expert updates are compatible and degrade when experts learn conflicting directions, which later methods address by resolving the conflict explicitly: TIES~\citep{yadav2023ties} by trimming and sign voting, DARE~\citep{yu2024dare} by random dropping, and KnOTS~\citep{stoica2024knots} in spectral space. A separate line weights parameters by importance, using curvature~\citep{matena2022fisher}, feature covariance~\citep{jin2023regmean}, or learned coefficients~\citep{yang2024adammerging}. All of them select by the size of a weight or singular value.

Our criterion follows Optimal Brain Damage~\citep{lecun1989obd} and the pruning results that succeeded it, where movement pruning~\citep{sanh2020movement} showed magnitude to be a particularly poor saliency signal in the fine-tuning regime we work in. Applying that idea to merging, \citet{park2026saliency} score task vectors with connectivity-based saliency and also decompose LoRAs rank-wise, and \citet{yang2025impart} vary sparsity per singular vector by importance instead of truncating by singular value. We differ in evaluating utility with a first-order gradient taken at the merged model rather than at individual experts, which is valid because the merged model sits at no single task's minimum, and in using the criterion to ask whether experts must share a rank at all.

Rank heterogeneity has been studied in adjacent settings, though for different reasons. Federated methods aggregate adapters of unequal rank because client hardware forces it~\citep{waseem2026prelort}, and adaptive methods allocate rank across layers within one adapter~\citep{zhang2023adalora}. Neither treats rank as a per-expert choice interacting with the merge operator. Closest to us, \citet{he2026compress} also identify the $R=\max_i r_i$ bottleneck and respond by compressing before merging, where we keep the budget and change which components fill it. LoRA~\citep{hu2022lora} is agnostic to rank, but merging is not. MergeBench~\citep{he2025mergebench} benchmarks merging methods and our setup follows its protocol, while concurrent work~\citep{kozodoi2026merging} finds expert training duration matters too, so both choices should be made jointly.

\section{Methods}
\label{sec:setup}

We merge with six operators spanning the two selection strategies at issue. Simple averaging and greedy soup~\citep{wortsman2022soups} combine weights without selecting components, the latter adding experts only when validation accuracy improves. TIES~\citep{yadav2023ties} trims low-magnitude entries and resolves sign conflicts, KnOTS~\citep{stoica2024knots} truncates the SVD of the stacked update, and plain SVD truncation keeps the top-$R$ singular directions. The last three all rank components by singular value, so they are magnitude-based.

Against these we use a gradient-guided selection method (GGSM), which keeps the same truncation budget but changes the criterion. Stacking the expert updates and taking the SVD gives components $(\sigma_j, u_j, v_j)$. We score each by projecting the combined task gradient $G$ onto it,
\begin{equation}
  \text{utility}_j = -\sigma_j\,(u_j^\top G v_j),
  \label{eq:utility}
\end{equation}
and keep the $R$ components of highest utility rather than largest $\sigma_j$. Positive utility means the component reduces loss to first order. To validate this criterion empirically, we sample 50 random components from a merged 0.8B model and measure the actual loss change from adding each. The predicted utility correlates with ground-truth loss change (Spearman $\rho = 0.49$, $p < 0.001$), outperforming magnitude ($\rho = 0.45$). However, the first-order approximation can overestimate utility for some components, particularly those with extreme predicted scores, suggesting room for higher-order corrections in future work. To compute $G$, we first merge with plain SVD (top-$\sigma$), then compute per-domain gradients $G_t = \partial \mathcal{L}_t / \partial W$ at the merged model using 64 validation samples per domain. Each gradient is normalized to unit Frobenius norm ($\hat{G}_t = G_t / \|G_t\|_F$) and summed with equal weight: $G = \sum_t \hat{G}_t$. This merged-point approximation is more accurate than evaluating at the base model because the first-order Taylor expansion is valid locally. Obtaining $G$ takes one backward pass per domain, so unlike the data-free baselines GGSM needs data access, at negligible cost.

We use the Qwen3.5 dense checkpoints at three scales, 0.8B, 2B and 4B, following the Qwen3 training recipe~\citep{qwen2025qwen3}. All experts are fine-tuned with QLoRA~\citep{dettmers2024qlora} on a 4-bit quantized
base, targeting all attention and MLP projection modules. We train with learning rate $2 \times 10^{-4}$, 100-step linear warmup, effective batch size 8, and sequence length 2048, selecting checkpoints by minimum validation loss. Rank is the only factor we vary: each domain is trained at ranks 4, 8, 16, and 32, giving 36 experts. Merges use a single seed, so we treat small score differences as ties throughout.

Following MergeBench~\citep{he2025mergebench}, we adopt three task domains. Mathematics uses DART-Math~\citep{wei2024dartmath} for training and GSM8K~\citep{cobbe2021gsm8k} for evaluation. Instruction following uses TULU-3~\citep{lambert2024tulu3} for training and IFEval~\citep{zhou2023ifeval} for evaluation. Multilingual uses Aya~\citep{singh2024aya} for training and reports the mean of MMLU~\citep{hendrycks2021mmlu}, ARC~\citep{clark2018arc}, and MMLU-zh for evaluation. Each domain has its own optimal rank for single-task performance, and these optima differ across domains and scales (Table~\ref{tab:single-task-opt}), so any uniform choice is suboptimal for at least one domain.

\begin{table}[t]
  \caption{Single-task optimal LoRA rank per domain and model. Optima differ across domains and
  shift with scale, so any uniform choice is suboptimal for at least one domain. Note that within each domain the variance across ranks is modest (typically $\leq$2 pp), so while optimum differs, the penalty for mismatched rank is small for single-task use; the larger effect comes from merging.}
  \label{tab:single-task-opt}
  \centering
  \begin{tabularx}{\linewidth}{lRRR}
    \toprule
    Model & Mathematics & Instruction & Multilingual \\
    \midrule
    Qwen3.5-0.8B & 8, 16, 32 & 16 & 4 \\
    Qwen3.5-2B   & 32        & 8  & 4 \\
    Qwen3.5-4B   & 4         & 32 & 8 \\
    \bottomrule
  \end{tabularx}
\end{table}

\section{Results}
\label{sec:results}

\subsection{Uniform ranks minimize what survives truncation}
\label{sec:structure}

\begin{table}[b]
  \caption{Spectral structure of the merged update (0.8B, averaged over 20--64 modules, two
independent samples agreeing to $\sim$2 decimals). \textsc{pool} is the mean numerical rank of the merged $\Delta W$ (count of singular values $> 10^{-6} \sigma_{\max}$, which is approximately $\sum_i r_i$ but varies slightly across modules due to near-zero components). $R=\max_i r_i$ is the truncation budget; \textsc{kept} $=R/\text{pool}$; and
\textsc{E@R} is the Frobenius energy retained by the top-$R$ components. Uniform assignments
sit at $\nicefrac{1}{3}$ kept at every rank, while diverse assignments retain
substantially more.}
  \label{tab:structure}
  \centering
  \begin{tabularx}{\linewidth}{llRRRR}
    \toprule
    Regime & Ranks (m/i/l) & $R$ & pool & kept & E@R \\
    \midrule
    uniform & 4/4/4       & 4  & 12.3 & 0.326 & 0.794 \\
    uniform & 8/8/8       & 8  & 24.1 & 0.332 & 0.827 \\
    uniform & 16/16/16    & 16 & 48.2 & 0.332 & 0.869 \\
    uniform & 32/32/32    & 32 & 96.2 & 0.332 & 0.876 \\
    diverse & 8/16/4      & 16 & 28.4 & 0.565 & 0.898 \\
    diverse & 4/16/32     & 32 & 52.4 & 0.611 & 0.891 \\
    diverse & 4/32/8      & 32 & 44.2 & \textbf{0.725} & \textbf{0.966} \\
    \bottomrule
  \end{tabularx}
\end{table}

Uniform ranks put the merge operator in the worst position it can occupy, and raising the rank does not help. With $T$ experts at a shared rank $r$, the pool holds $\approx Tr$ components while the budget admits $R=r$, so a fixed $1/T$ survives however much capacity each expert was given. Table~\ref{tab:structure} shows this pinned near one-third across an $8\times$ range of ranks. Diverse ranks break the pattern, because the budget follows the largest expert while the pool follows the sum: assigning $r=4$ to math, $32$ to instruction and $8$ to multilingual leaves $\approx 44$ components with $R=32$, so 73\% survive. A shared rank costs no capacity inside any expert, only the share of the combined update that survives.

This bottleneck applies only to methods that truncate the merged spectrum (SVD, KnOTS, TIES, and GGSM). Simple averaging keeps the full sum $\sum_i \Delta W_i$ without truncation, so rank heterogeneity is free by construction---there is no $R$ budget to exhaust. The empirical results in Section~\ref{sec:forcing} confirm this: averaging tolerates diverse ranks as well as GGSM does, even though GGSM actively selects components while averaging does not.

\begin{table}[t]
  \caption{Utility versus magnitude selection at identical budget $R$ (0.8B, base-point
  gradients, 96 target matrices). \textsc{frac.\ pos.}\ exceeds \textsc{kept} in every
  configuration: the budget must reject loss-reducing directions. \textsc{clipped} is the mean
  count per matrix that utility keeps and magnitude discards.}
  \label{tab:overlap}
  \centering
  \begin{tabularx}{\linewidth}{llRRRRRR}
    \toprule
    Regime & Ranks & kept & overlap & clipped & frac.\ pos. & $E_\sigma$ & $E_{\mathrm{util}}$ \\
    \midrule
    uniform & 4/4/4    & 0.333 & 0.747 & 1.01 & 0.795 & 0.797 & 0.694 \\
    uniform & 8/8/8    & 0.333 & 0.714 & 2.29 & 0.755 & 0.827 & 0.686 \\
    uniform & 16/16/16 & 0.333 & 0.669 & 5.30 & 0.720 & 0.866 & 0.665 \\
    uniform & 32/32/32 & 0.333 & 0.694 & 9.78 & 0.652 & 0.876 & 0.701 \\
    diverse & 8/16/4   & 0.571 & 0.744 & 4.09 & \textbf{0.843} & 0.896 & 0.783 \\
    diverse & 4/16/32  & 0.615 & 0.687 & 10.02 & 0.780 & 0.892 & 0.718 \\
    \bottomrule
  \end{tabularx}
\end{table}

Equation~\ref{eq:utility} lets us ask whether the discarded components were worth keeping, and by that measure the budget is not merely tight, it is spent on the wrong components. Table~\ref{tab:overlap} shows that 65 to 84\% of pooled components carry positive utility while the budget admits 33\% under uniform ranks and at most 62\% under diverse ones, so useful directions must be dropped whatever the hyperparameters, and the waste grows with rank: the count of high-utility components magnitude keeps out rises from about 1 to nearly 10 per matrix between $r=4$ and $r=32$, tracking the decline of TIES while utility-based selection stays flat (Figure~\ref{fig:discard}). Utility selection retains less spectral energy ($E_{\mathrm{util}} < E_\sigma$) and still merges better, so what matters is which components survive, not how much energy they carry.

\subsection{Magnitude methods lose more on diverse ranks}
\label{sec:forcing}

\begin{table}[t]
  \caption{Method comparison across scales, best diverse configuration per method, where
  $\Delta = (\text{best diverse}) - (\text{best uniform})$. GGSM, SVD and averaging hold within
  noise, TIES and KnOTS perform inconsistently across scales while greedy soup degrades. IFEval drives the spread.}
  \label{tab:forcing}
  \centering
  \begin{tabularx}{\linewidth}{llRRRRR}
    \toprule
    Scale & Method & Math & IFEval & Multi & Avg & $\Delta$ \\
    \midrule
    \multirow{6}{*}{0.8B}
    & GGSM        & 40.9 & \textbf{55.1} & 50.7 & \textbf{48.9} & $+1.4$ \\
    & SVD         & 42.4 & 48.6 & 51.2 & 47.4 & $+0.3$ \\
    & Simple avg  & 44.9 & 49.7 & 50.9 & 48.5 & $+1.8$ \\
    & Greedy soup & 45.3 & 39.9 & 48.0 & 44.4 & $\mathbf{-4.1}$ \\
    & TIES        & 37.6 & 39.7 & 47.9 & 41.7 & $-0.5$ \\
    & KnOTS       & 37.9 & 34.2 & 47.6 & 39.9 & $-2.1$ \\
    \midrule
    \multirow{6}{*}{2B}
    & GGSM        & 61.5 & \textbf{67.5} & 58.2 & \textbf{62.4} & $+0.7$ \\
     & SVD         & 64.1 & 64.5 & 58.3 & 62.3 & $+0.2$ \\
    & Simple avg  & 62.5 & 63.4 & 58.3 & 61.4 & $+0.3$ \\
    & Greedy soup & 64.5 & 61.2 & 55.1 & 60.3 & $-1.0$ \\
    & TIES        & 62.8 & 54.3 & 55.9 & 57.7 & $+0.4$ \\
    & KnOTS       & 62.9 & 51.9 & 52.8 & 55.9 & $+2.3$ \\
    \midrule
    \multirow{6}{*}{4B}
    & GGSM        & 82.5 & \textbf{82.3} & 70.4 & \textbf{78.4} & $+0.4$ \\
    & SVD         & 84.2 & 78.7 & 70.4 & 77.8 & $-0.3$ \\
    & Simple avg  & 83.9 & 79.3 & 70.2 & 77.8 & $-0.1$ \\
    & Greedy soup & 83.8 & 78.9 & 70.4 & 77.7 & $0.0$ \\
    & TIES        & 83.8 & 61.2 & 68.2 & 71.1 & $\mathbf{-4.0}$ \\
    & KnOTS       & 82.6 & 54.0 & 68.5 & 68.4 & $-1.1$ \\
    \bottomrule
  \end{tabularx}
\end{table}

If the structural argument holds, methods should divide by how they choose components rather than by how sophisticated they are, and Table~\ref{tab:forcing} shows exactly that. Moving each method from its best uniform assignment to its best diverse one separates the six operators into two groups. For GGSM, SVD and Simple averaging every delta lies between $-0.3$ and $+1.8$ pp, inside our interval, and the direction is not random: GGSM improves at all three scales by 0.4 to 1.4 pp, and the largest single gain is 1.8 pp for Simple averaging at 0.8B. Diverse ranks are free for these three, with a hint of a gain our sample sizes cannot resolve. Greedy soup loses ground at 0.8B and 2B and is flat at 4B, while TIES and KnOTS are inconsistent; their best diverse configurations beat uniform at 2B but trail at 0.8B and 4B.

\begin{figure}[b]
  \centering
  \includegraphics[width=0.48\linewidth]{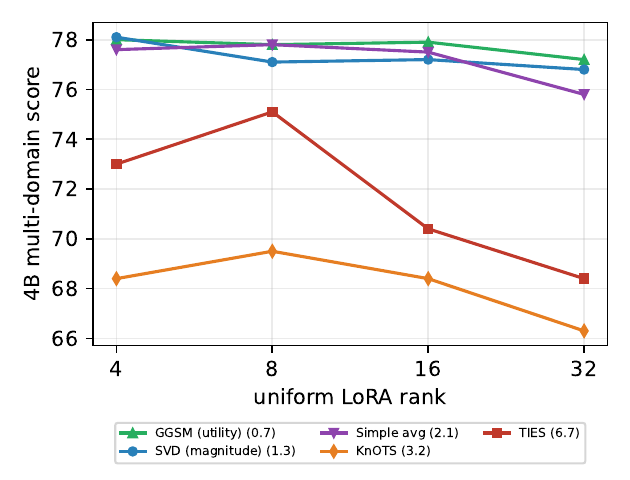}
  \hfill
  \includegraphics[width=0.48\linewidth]{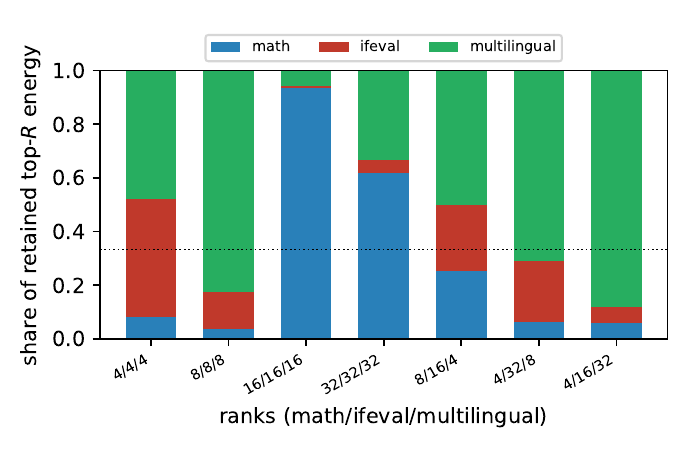}
  \caption{\textbf{Left:} sensitivity to uniform rank at 4B, where utility selection spans 0.7 pp against 6.7 pp for TIES. \textbf{Right:} share of the retained top-$R$ subspace per expert under magnitude truncation (SVD), where
  one expert is often dropped almost entirely (e.g.\ $16/16/16 \to 0.93/0.01/0.06$). GGSM selects by utility rather than magnitude, so the eviction pattern differs; we leave that analysis to future work.}
  \label{fig:sensitivity}
\end{figure}

Rank sensitivity tells the same story from a different angle. Across the uniform-rank sweep at 4B, GGSM spans 0.7 pp while TIES spans 6.7, about $9\times$ more (Figure~\ref{fig:sensitivity}, left). A method that ignores utility does not merely mishandle diverse ranks, it is unstable under the choice of rank in general. The ordering (GGSM $<$ SVD $<$ averaging $<$ KnOTS $<$ TIES) tracks reliance on magnitude. Appendix~\ref{app:sensitivity} gives the full sweep, and Appendix~\ref{app:diverse} shows that at 4B, every diverse configuration costs TIES 4 to 9 pp, at 0.8B under 1 pp, and only at 2B does the best diverse setup edge ahead.

Which of these differences we can trust depends on a resolution floor, which we derive in Appendix~\ref{app:ci} as roughly 1.4 pp for the paired comparisons we report. Three results clear it: the TIES penalty of $-4$ pp, the sensitivity ranges of 6.7 against 0.7 pp, and the IFEval gaps of 16 to 25 pp. The gaps among the four leading methods---GGSM, SVD, averaging, and greedy soup---do not clear this floor, so we treat those as statistically tied. In particular, GGSM's advantage over SVD is small (0.6 pp at 4B, 0.1 pp at 2B, 1.5 pp at 0.8B) and within noise; the value of GGSM lies in its rank stability (0.7 pp range vs 1.3 pp for SVD at 4B) rather than in higher peak accuracy.

\subsection{Instruction following is suppressed}
\label{sec:eviction}

An average score hides the shape of the damage, which is concentrated rather than spread. Attributing the retained top-$R$ subspace to individual experts through $\langle \Delta W_d, C_j\rangle$ over kept components $j$ shows it is badly unbalanced: in 6 of 7 configurations one expert holds at least 0.64 of the retained energy while another is driven near zero, reaching $0.93/0.01/0.06$ at uniform rank 16 (Figure~\ref{fig:sensitivity}, right). Merging is meant to combine experts, but under magnitude truncation it can quietly delete one, a capability loss introduced by the merge rather than absent beforehand.

\begin{table}[t]
  \caption{Per-domain scores at uniform rank 4 at 0.8B, 2B and 4B. IFEval collapses for
  TIES and KnOTS at every scale.}
  \label{tab:perdomain-r4}
  \centering
  \begin{tabularx}{\linewidth}{llRRRR}
    \toprule
    Scale & Method & Math & IFEval & Multi & Avg \\
    \midrule
    \multirow{6}{*}{0.8B}
    & GGSM (utility)   & 38.6 & \textbf{55.5} & 48.5 & 47.5 \\
    & SVD (magnitude)  & 39.7 & 47.7 & 50.8 & 46.1 \\
    & Simple averaging & 42.7 & 47.1 & 50.3 & 46.7 \\
    & Greedy soup      & 44.7 & 50.8 & 50.0 & \textbf{48.5} \\
    & TIES             & 39.3 & 39.2 & 48.1 & 42.2 \\
    & KnOTS            & 41.9 & 36.0 & 48.2 & 42.1 \\
    \midrule
    \multirow{6}{*}{2B}
    & GGSM (utility)   & 58.6 & \textbf{67.1} & 56.5 & 60.7 \\
    & SVD (magnitude)  & 63.8 & 63.8 & 58.7 & \textbf{62.1} \\
    & Simple averaging & 62.2 & 63.4 & 57.3 & 60.9 \\
    & Greedy soup      & 64.6 & 61.6 & 57.6 & 61.3 \\
    & TIES             & 60.7 & 50.8 & 58.1 & 56.5 \\
    & KnOTS            & 57.1 & 42.1 & 55.7 & 51.6 \\
    \midrule
    \multirow{6}{*}{4B}
    & GGSM (utility)   & 81.7 & \textbf{82.3} & 70.0 & 78.0 \\
    & SVD (magnitude)  & 83.7 & 80.8 & 69.7 & \textbf{78.1} \\
    & Simple averaging & 84.1 & 78.7 & 69.9 & 77.6 \\
    & Greedy soup      & 84.0 & 79.1 & 70.0 & 77.7 \\
    & TIES             & 84.2 & 66.0 & 68.7 & 73.0 \\
    & KnOTS            & 80.9 & 56.6 & 67.7 & 68.4 \\
    \bottomrule
  \end{tabularx}
\end{table}

Table~\ref{tab:forcing} compares each method's best diverse-rank configuration against the single-task experts of Table~\ref{tab:experts-all}. On the Math task, every method exceeds its expert at all three scales, combining experts yields consistent positive transfer for Math capability. On Multilingual task, GGSM and SVD exceed the single-task experts at every scale while the remaining methods, TIES and KnOTS in particular, fall below. The Instruction following task is the discriminating axis: GGSM is the only method that surpasses the single-task IFEval expert at every scale (0.8B: 55.1 vs.\ 54.2; 2B: 67.5 vs.\ 67.3; 4B: 82.3 vs.\ 81.7), whereas all other methods trail it and TIES and KnOTS collapse (e.g.\ 2B: 54.3 and 51.9). GGSM is thus the only method to improve on the single-task expert across all three domains simultaneously, which is why it attains the highest average at every scale (48.9, 62.4, 78.4) and stands as the strongest method overall.

\section{Conclusion}
\label{sec:conclusion}

Established practice in model merging trains every expert at the same LoRA rank. Our results suggest this convention reflects a property of the merge operator rather than a requirement of merging. It comes specifically from conflict-resolution methods that select components by magnitude: TIES and KnOTS are sensitive to rank heterogeneity while SVD truncation is not, even though all three rank components by singular value. The difference is that TIES and KnOTS apply additional operations (trimming, sign election) that interact with the magnitude distribution in ways that break under heterogeneous ranks. With mismatched ranks, the low-rank expert has fewer but denser parameters compared to a high-rank expert with many but sparser parameters; when TIES trims by magnitude percentile and elects signs by majority vote, the low-rank expert's concentrated magnitudes get disproportionately trimmed while its smaller vote share loses sign elections. The combined effect pushes the low-rank expert out of the merged result. With averaging or utility-based selection, diverse ranks cost nothing and we measure gains of up to 1.8 pp, so rank becomes a per-expert decision that needs no coordination. Evaluated with TIES the same change costs 4 pp at 4B, and KnOTS is inconsistent. For TIES and KnOTS, the safer choice is matching ranks, but that conclusion does not generalize to merging or even to all magnitude-based methods. This matters most when experts come from separate teams or arrive at different times. The practical recommendation is to train each expert at the rank that suits its domain, and merge with an operator that is rank-insensitive---whether by keeping the full sum (averaging), by selecting components on effect rather than size (GGSM), or by truncating without conflict resolution (SVD).

Our study has limitations we plan to address in future work. The small gains from diverse ranks sit below our 1.4 pp resolution floor, so confirming their significance needs more seeds and larger evaluations. We also cover three domains, focus on LoRA fine-tuning only, and use one model family.

% flush body floats so appendix tables cannot drift up into the reference list
\FloatBarrier
\bibliographystyle{unsrtnat}
\bibliography{refs}

\clearpage
\appendix

% denser float packing in the appendix, where pages are mostly tables and figures
\setlength{\textfloatsep}{10pt plus 2pt minus 2pt}
\setlength{\floatsep}{8pt plus 2pt minus 2pt}
\setlength{\intextsep}{8pt plus 2pt minus 2pt}
\renewcommand{\topfraction}{0.95}
\renewcommand{\bottomfraction}{0.95}
\renewcommand{\textfraction}{0.05}
\renewcommand{\floatpagefraction}{0.6}
\setcounter{topnumber}{3}
\setcounter{bottomnumber}{2}
\setcounter{totalnumber}{5}

\begin{table}[!b]
  \caption{Uniform-rank sweep across scales (same-rank merges). GGSM is the most
  rank-stable at every scale; KnOTS and TIES are the most sensitive.}
  \label{tab:sweep-all}
  \centering
  \begin{tabularx}{\linewidth}{llRRRRR}
    \toprule
    Scale & Method & r4 & r8 & r16 & r32 & Range \\
    \midrule
    \multirow{5}{*}{0.8B}
    & GGSM (utility)   & 47.5 & 47.5 & 46.9 & 46.3 & \textbf{1.2} \\
    & SVD (magnitude)  & 46.1 & 46.8 & 47.1 & 45.9 & 1.2 \\
    & Simple averaging & 46.7 & 46.5 & 46.2 & 44.9 & 1.8 \\
    & KnOTS            & 42.1 & 37.7 & 35.0 & 34.6 & 7.5 \\
    & TIES             & 42.2 & 41.0 & 39.7 & 37.9 & 4.3 \\
    \midrule
    \multirow{5}{*}{2B}
    & GGSM (utility)   & 60.7 & 61.2 & 61.7 & 60.6 & \textbf{1.1} \\
    & SVD (magnitude)  & 62.1 & 61.0 & 61.9 & 59.8 & 2.3 \\
    & Simple averaging & 60.9 & 61.1 & 60.6 & 58.5 & 2.6 \\
    & KnOTS            & 51.6 & 53.6 & 52.2 & 51.0 & 2.6 \\
    & TIES             & 56.5 & 57.3 & 55.5 & 53.6 & 3.7 \\
    \midrule
    \multirow{5}{*}{4B}
    & GGSM (utility)   & 78.0 & 77.8 & 77.9 & 77.2 & \textbf{0.7} \\
    & SVD (magnitude)  & 78.1 & 77.1 & 77.2 & 76.8 & 1.3 \\
    & Simple averaging & 77.6 & 77.8 & 77.5 & 75.8 & 2.1 \\
    & KnOTS            & 68.4 & 69.5 & 68.4 & 66.3 & 3.2 \\
    & TIES             & 73.0 & 75.1 & 70.4 & 68.4 & 6.7 \\
    \bottomrule
  \end{tabularx}
\end{table}

\begin{figure}[!b]
  \centering
  \includegraphics[width=0.48\linewidth]{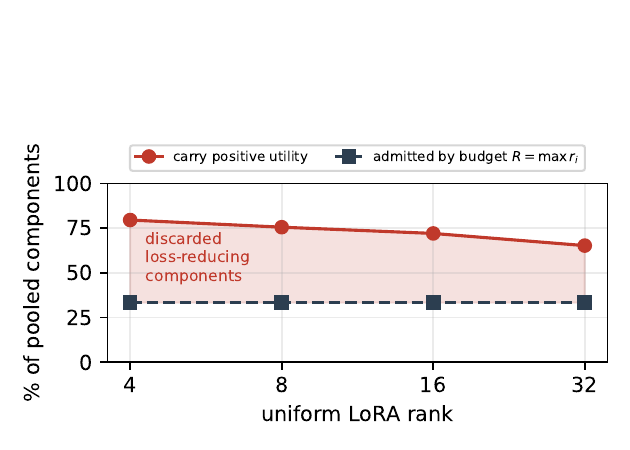}
  \hfill
  \includegraphics[width=0.48\linewidth]{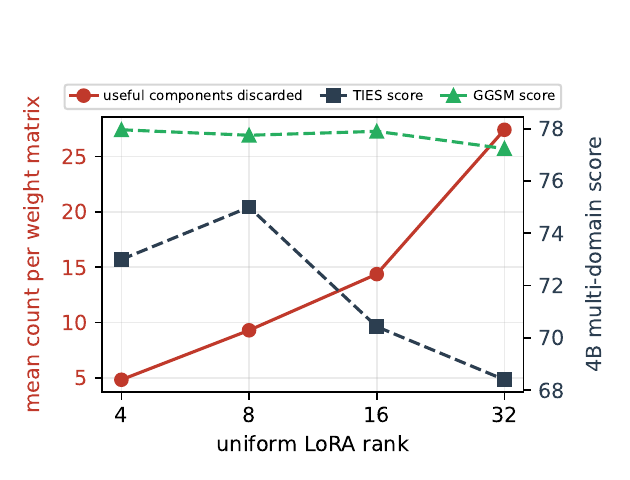}
  \caption{\textbf{Left:} the fraction of pooled components carrying positive first-order utility far exceeds the fraction the budget admits, which is fixed at $\frac{1}{T}$ for $T$ tasks under
  any uniform assignment. \textbf{Right:} the number of high-utility components that magnitude selection discards grows roughly tenfold from $r=4$ to $r=32$ (left axis), tracking the decline of TIES while utility-based selection stays flat (right axis).}
  \label{fig:discard}
\end{figure}

\section{Resolution Floor for Score Comparisons}
\label{app:ci}

Per-benchmark standard errors propagate as $\mathrm{SE}_{\text{domain}}=\sqrt{\sum_i \mathrm{se}_i^2}/k$ over the $k$ benchmarks in a domain, and $\mathrm{SE}_{\text{avg}}=\sqrt{\sum_d \mathrm{SE}_d^2}/3$ over the three domains. This gives $\mathrm{SE}_{\text{avg}}\approx0.93$ to $1.04$ pp, so a $95\%$ interval spans roughly $\pm1.9$ pp on any single multi-domain score. The diverse-versus-uniform comparisons are paired, since both configurations are evaluated on identical test items, and a paired bootstrap over those shared items roughly halves the interval. We therefore treat 1.4 pp as the smallest difference we can resolve, and read anything below it as a tie.

\section{How Much the Merge Method Depends on Rank}
\label{app:sensitivity}

Table~\ref{tab:sweep-all} gives the full uniform-rank sweep at 0.8B, 2B and 4B Qwen3.5 models. At every scale GGSM has the most stable performance across 8× range while TIES drops by 4.3 at 0.8B scale and by 6.7 at 4B scale. The ordering by range (GGSM $<$ SVD $<$ averaging $<$ KnOTS $<$ TIES) matches reliance on magnitude-based selection. Figure~\ref{fig:discard} shows the mechanism behind that ordering: the share of pooled components carrying positive utility far exceeds what the budget admits, and the number of useful components discarded grows with rank.

\begin{table}[!t]
  \caption{TIES on diverse ranks across scales. At every size the best diverse configuration underperforms the best uniform-rank baseline (0.8B: 42.2 at r4; 2B: 57.3 at r8; 4B: 75.1 at r8) except at 2B where the diverse-rank setup is marginally higher.}
  \label{tab:ties-diverse-all}
  \centering
  \begin{tabularx}{\linewidth}{llRRRR}
    \toprule
    Scale & Ranks (m/i/l) & Math & IFEval & Multi & Avg \\
    \midrule
    \multirow{5}{*}{0.8B}
    & 8/32/4 (optimal) & 37.6 & 39.7 & 47.9 & 41.7 \\
    & 4/8/16           & 39.3 & 39.4 & 46.2 & 41.6 \\
    & 8/16/4           & 39.0 & 38.4 & 46.9 & 41.4 \\
    & 4/32/8           & 35.6 & 38.4 & 50.1 & 41.4 \\
    & 32/8/4           & 39.9 & 35.3 & 48.8 & 41.3 \\
    \midrule
    \multirow{5}{*}{2B}
    & 16/8/4 (optimal) & 62.8 & 54.3 & 55.9 & \textbf{57.7} \\
    & 16/4/8           & 61.9 & 54.3 & 55.8 & 57.3 \\
    & 16/32/8          & 60.9 & 52.3 & 56.7 & 56.6 \\
    & 32/16/4          & 61.7 & 53.4 & 54.5 & 56.5 \\
    & 32/8/4           & 61.5 & 53.4 & 54.3 & 56.4 \\
    \midrule
    \multirow{5}{*}{4B}
    & 4/32/8 (optimal) & 83.8 & 61.2 & 68.2 & 71.1 \\
    & 4/16/8           & 82.9 & 56.0 & 68.1 & 69.0 \\
    & 4/8/16           & 81.4 & 57.5 & 68.1 & 69.0 \\
    & 4/16/32          & 80.7 & 51.6 & 67.7 & 66.7 \\
    & 4/8/32           & 81.0 & 50.1 & 67.4 & 66.1 \\
    \bottomrule
  \end{tabularx}
\end{table}

\section{The Cost of Diverse Ranks}
\label{app:diverse}

Table~\ref{tab:ties-diverse-all} reports TIES across diverse-rank configurations at all three model scales, from which two observations follow. First, the natural heuristic of assigning each domain its individually optimal rank (0.8B: 8/16/4; 2B: 32/8/4; 4B: 4/32/8) yields no improvement: it underperforms the best uniform-rank baseline at every scale (0.8B: 41.4 vs.\ 42.2, $-0.8$; 2B: 56.4 vs.\ 57.3, $-0.9$; 4B: 71.1 vs.\ 75.1, $-4.0$). Second, an exhaustive search over diverse assignments does recover a configuration that matches or marginally exceeds the uniform baseline at the two smaller scales (best diverse at 0.8B, 41.7, $-0.5$; at 2B, 57.7, $+0.4$), whereas at 4B even the best diverse configuration falls short by 4.0~pp. For TIES, therefore, the benefit of rank diversity is marginal and attainable only through search; the per-domain-optimal heuristic is consistently inferior to a single shared rank, and this deficit grows with scale. Where degradation occurs, it is attributable to IFEval, which deteriorates sharply under diverse-rank TIES.

Table~\ref{tab:diverse-all} extends this comparison to all methods evaluated at the per-domain-optimal ranks across scales. The utility-, magnitude-, and averaging-based methods (GGSM, SVD, simple averaging, and Greedy soup) lie within approximately 1pp in average score at 2B (60.3-61.3) and 4B (77.7-78.4), inside our interval, so we treat them as tied, while GGSM establishes a wider margin of roughly 2pp at 0.8B (48.5 vs.\ 46.6). TIES and KnOTS trail this group by 4 to 10pp at every scale, again owing to the collapse of IFEval: GGSM preserves IFEval performance at 56.7, 66.7, and 82.3 for 0.8B, 2B, and 4B respectively, whereas TIES falls to 38.4, 53.4, and 61.2. This pattern is thus consistent across scales rather than specific to any single model size.

\begin{table}[!t]
  \caption{All methods on the per-domain optimal ranks at each scale
  (0.8B: 8/16/4; 2B: 32/8/4; 4B: 4/32/8). GGSM leads on Avg at every scale;
  TIES and KnOTS trail because IFEval collapses.}
  \label{tab:diverse-all}
  \centering
  \begin{tabularx}{\linewidth}{llRRRR}
    \toprule
    Scale & Method & Math & IFEval & Multi & Avg \\
    \midrule
    \multirow{6}{*}{0.8B}
    & GGSM (utility)   & 39.0 & 56.7 & 49.8 & \textbf{48.5} \\
    & Simple averaging & 41.6 & 48.2 & 49.8 & 46.5 \\
    & SVD (magnitude)  & 41.3 & 47.3 & 51.2 & 46.6 \\
    & Greedy soup      & 45.3 & 39.9 & 48.0 & 44.4 \\
    & KnOTS            & 37.9 & 34.2 & 47.6 & 39.9 \\
    & TIES             & 39.0 & 38.4 & 46.9 & 41.4 \\
    \midrule
    \multirow{6}{*}{2B}
    & GGSM (utility)   & 61.9 & 66.7 & 55.2 & \textbf{61.3} \\
    & Simple averaging & 63.5 & 63.6 & 55.9 & 61.0 \\
    & SVD (magnitude)  & 64.4 & 61.0 & 55.8 & 60.4 \\
    & Greedy soup      & 64.5 & 61.2 & 55.1 & 60.3 \\
    & KnOTS            & 62.9 & 51.9 & 52.8 & 55.9 \\
    & TIES             & 61.5 & 53.4 & 54.3 & 56.4 \\
    \midrule
    \multirow{6}{*}{4B}
    & GGSM (utility)   & 82.5 & 82.3 & 70.4 & \textbf{78.4} \\
    & Simple averaging & 83.8 & 79.3 & 70.2 & 77.8 \\
    & SVD (magnitude)  & 84.2 & 78.7 & 70.4 & 77.8 \\
    & Greedy soup      & 83.8 & 78.9 & 70.4 & 77.7 \\
    & KnOTS            & 82.6 & 54.0 & 68.5 & 68.4 \\
    & TIES             & 83.8 & 61.2 & 68.2 & 71.1 \\
    \bottomrule
  \end{tabularx}
\end{table}

\begin{table}[!t]
  \caption{2B results: Greedy soup vs GGSM on same-rank and diverse-rank configurations.}
  \label{tab:2b-results}
  \centering
  \begin{tabularx}{\linewidth}{llRRRR}
    \toprule
    Method & Ingredients & Math & IFEval & Multi & Avg \\
    \midrule
    GGSM (same)    & 16/16/16          & 61.4 & \textbf{67.1} & 56.7 & \textbf{61.7} \\
    Greedy (same)  & m4, i4            & 64.6 & 61.6 & 57.6 & 61.3 \\
    GGSM (diverse) & 8/16/4            & 61.5 & 67.5 & 58.2 & 62.4 \\
    Greedy (diverse) & m32, i4, l8     & 64.5 & 61.2 & 55.1 & 60.3 \\
    \bottomrule
  \end{tabularx}
\end{table}

\vspace{1em}
\section{Where the Loss Lands}
\label{app:collapse}

\begin{figure}[!b]
  \centering
  \includegraphics[width=\linewidth]{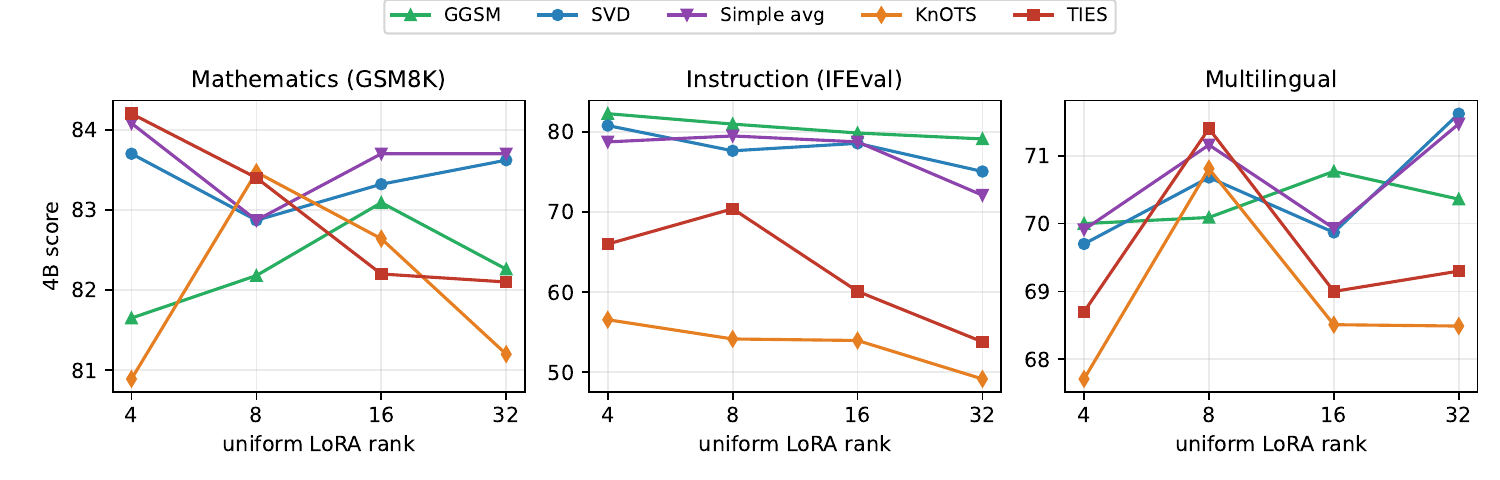}
  \caption{Per-domain scores across uniform ranks at 4B. IFEval (center) shows the starkest
  method divergence: GGSM and SVD maintain 75.1 to 82.3 while TIES and KnOTS collapse to 49.2 to 70.4.
  Math (left) and Multilingual (right) vary little across methods.}
  \label{fig:perdomain}
\end{figure}

\begin{figure}[!b]
  \centering
  \includegraphics[width=0.48\linewidth]{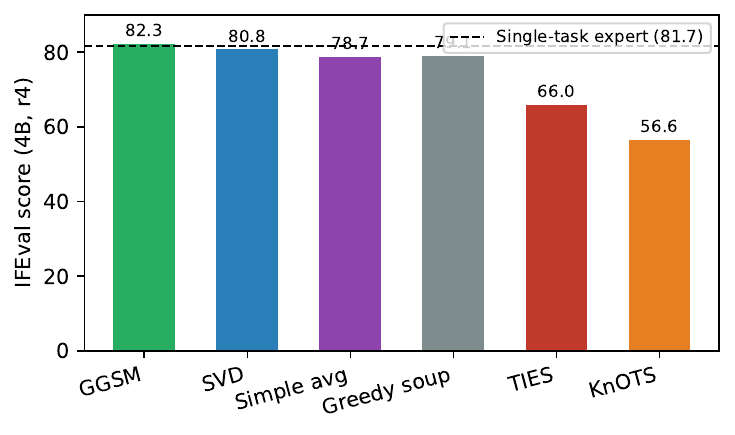}
  \hfill
  \includegraphics[width=0.48\linewidth]{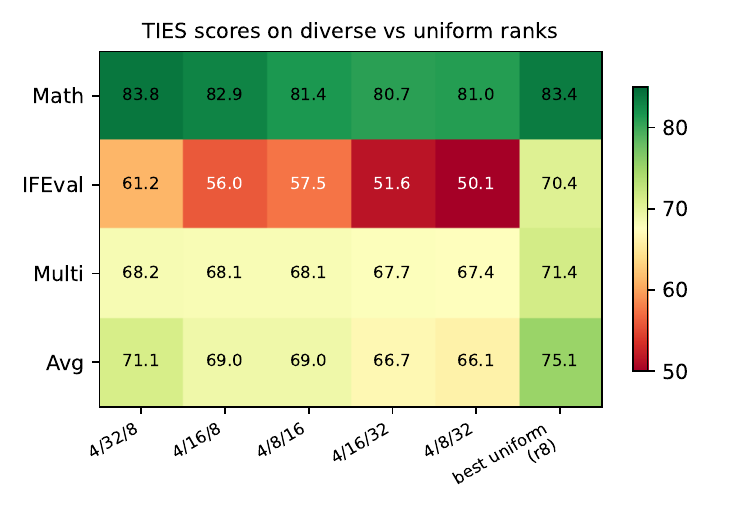}
  \caption{\textbf{Left:} IFEval scores at 4B, uniform rank 4. TIES and KnOTS lose 16 to 25 pp
  relative to the single-task expert baseline (dashed line), while GGSM slightly exceeds it.
  \textbf{Right:} TIES performance across diverse rank configurations vs best uniform (r8).
  Every diverse configuration underperforms uniform on aggregate, primarily due to IFEval collapse.}
  \label{fig:ifeval-and-ties}
\end{figure}

Table~\ref{tab:experts-all} lists the single-task baselines the merged models are measured against. Merged models can exceed a single-task expert on its own domain through positive transfer---for instance, stronger instruction following may improve how math answers are formatted and thereby extracted. Figure~\ref{fig:perdomain} breaks the uniform-rank sweep down by domain and shows that the method spread is dominated by instruction following (at 4B, a 33-point range across methods versus 3.3 for math and 3.9 for multilingual), while Figure~\ref{fig:ifeval-and-ties} compares IFEval against the single-task baseline and sweeps TIES across diverse configurations.

\begin{table}[!t]
  \caption{Single-task expert baselines across scales (best rank per domain and its score).}
  \label{tab:experts-all}
  \centering
  \begin{tabularx}{\linewidth}{llRR}
    \toprule
    Scale & Domain & Best Rank & Score \\
    \midrule
    \multirow{3}{*}{0.8B}
    & Math (GSM8K)         & r8/r16/r32  & 34.5 \\
    & Instruction (IFEval) & r16 & 54.2 \\
    & Multilingual         & r4  & 48.0 \\
    \midrule
    \multirow{3}{*}{2B}
    & Math (GSM8K)         & r32 & 56.6 \\
    & Instruction (IFEval) & r8  & 67.3 \\
    & Multilingual         & r4  & 57.7 \\
    \midrule
    \multirow{3}{*}{4B}
    & Math (GSM8K)         & r4  & 79.7 \\
    & Instruction (IFEval) & r32 & 81.7 \\
    & Multilingual         & r8  & 70.3 \\
    \bottomrule
  \end{tabularx}
\end{table}


\end{document}